\documentclass[10pt,twocolumn,letterpaper]{article}

\usepackage{cvpr}              
\definecolor{cvprblue}{rgb}{0.21,0.49,0.74}
\usepackage[pagebackref,breaklinks,colorlinks,allcolors=cvprblue]{hyperref}
\usepackage{adjustbox}
\usepackage[table]{xcolor}
\usepackage{amsmath}
\usepackage{multirow}
\usepackage{booktabs}
\usepackage{pifont}
\usepackage[accsupp]{axessibility}

\def\paperID{12835} 
\def\confName{CVPR}
\def\confYear{2026}

\title{MoE-GRPO: Optimizing Mixture-of-Experts via \\ Reinforcement Learning in Vision-Language Models}

\author{
Dohwan Ko\textsuperscript{\rm 1}\hspace{0.2cm}
Jinyoung Park\textsuperscript{\rm 2}\hspace{0.2cm}
Seoung Choi\textsuperscript{\rm 2}\hspace{0.2cm}
Sanghyeok Lee\textsuperscript{\rm 2}\hspace{0.2cm}
Seohyun Lee\textsuperscript{\rm 2}\hspace{0.2cm}
Hyunwoo J. Kim\textsuperscript{\rm 2}$^{\dagger}$\vspace{0.3cm} \\
\textsuperscript{\rm 1}Korea University\hspace{0.8cm}
\textsuperscript{\rm 2}KAIST\vspace{0.3cm} \\
\tt\small ikodoh@korea.ac.kr \\
\tt\small \{jinyoung.park, choisw0823, sanghyeoklee, seohyunlee, hyunwoojkim\}@kaist.ac.kr
}

\begin{document}
\maketitle

\renewcommand{\thefootnote}{\fnsymbol{footnote}}
\footnotetext[0]{$\dagger$ Corresponding authors.}

\begin{abstract}
    Mixture-of-Experts (MoE) has emerged as an effective approach to reduce the computational overhead of Transformer architectures by sparsely activating a subset of parameters for each token while preserving high model capacity.
    This paradigm has recently been extended to Vision-Language Models (VLMs), enabling scalable multi-modal understanding with reduced computational cost.
    However, the widely adopted deterministic top-$K$ routing mechanism may overlook more optimal expert combinations and lead to expert overfitting.
    To address this limitation and improve the diversity of expert selection, we propose \textbf{MoE-GRPO}, a reinforcement learning~(RL)-based framework for optimizing expert routing in MoE-based VLMs.
    Specifically, we formulate expert selection as a sequential decision-making problem and optimize it using Group Relative Policy Optimization (GRPO), allowing the model to learn adaptive expert routing policies through exploration and reward-based feedback.
    Furthermore, we introduce a \textbf{modality-aware router guidance} that enhances training stability and efficiency by discouraging the router from exploring experts that are infrequently activated for a given modality.
    Extensive experiments on multi-modal image and video benchmarks show that MoE-GRPO consistently outperforms standard top-$K$ routing and its variants by promoting more diverse expert selection, thereby mitigating expert overfitting and enabling a task-level expert specialization.
\end{abstract}
\section{Introduction}

\begin{figure}[t]
    \begin{subfigure}[t]{0.49\linewidth}
         \centering
         \includegraphics[width=\linewidth]{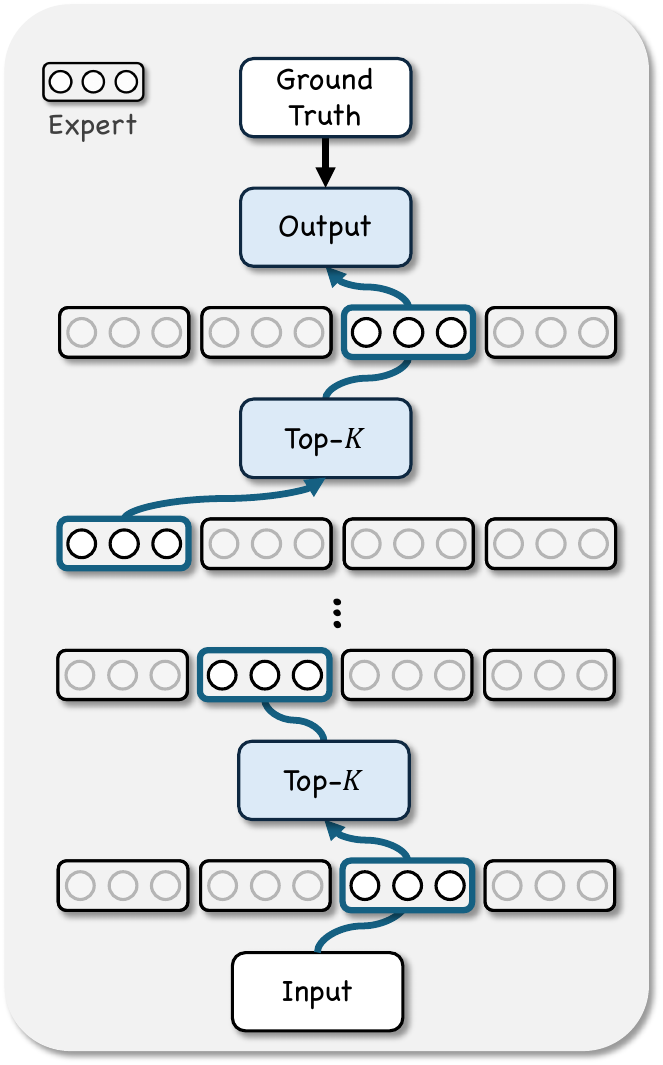}
         \caption{\textbf{Top-$K$ routing.}}
         \label{fig:teaser1}
    \end{subfigure}
    \begin{subfigure}[t]{0.49\linewidth}
         \centering
         \includegraphics[width=\linewidth]{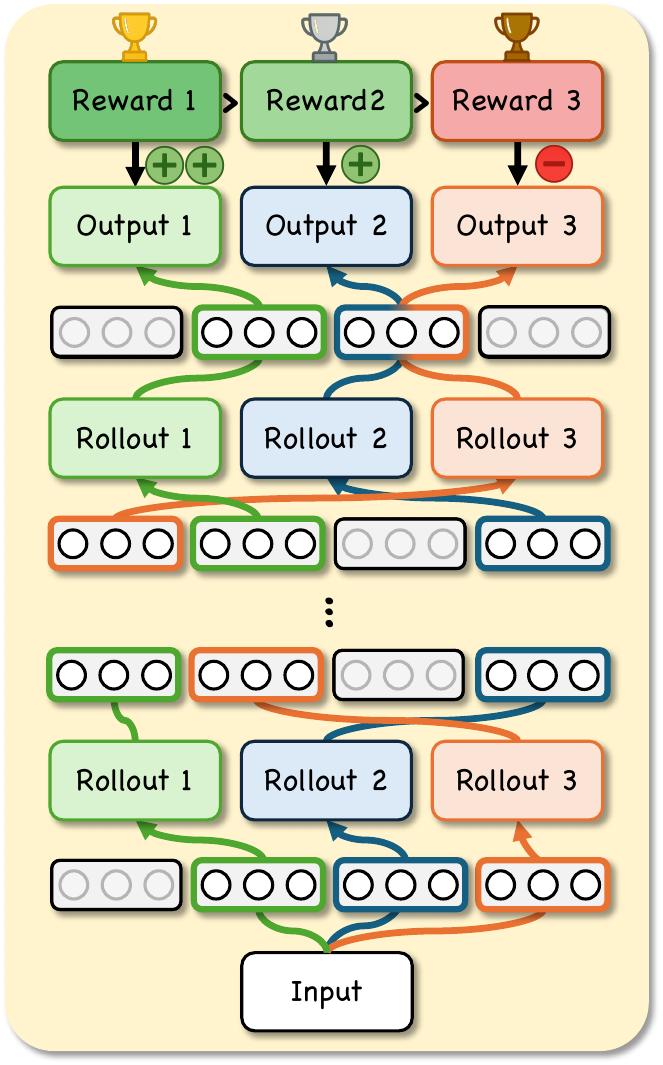}
         \caption{\textbf{MoE-GRPO.}}
         \label{fig:teaser2}
    \end{subfigure}
    \caption{
        \textbf{Comparison between top-$K$ routing and our MoE-GRPO.}
        (a) While the top-$K$ routing deterministically selects $K$ experts based on gating scores, (b) MoE-GRPO stochastically samples $K$ experts across multiple rollouts and optimizes the expert selection policy through reward-based feedback.
    }
    \label{fig:teaser}
    \vspace{-1mm}
\end{figure}

Scaling model capacity, particularly in Transformer architectures~\cite{vaswani2017attention}, has led to unprecedented performance gains in Large Language Models (LLMs)~\cite{touvron2023llama,achiam2023gpt,yang2025qwen3,cai2024internlm2,team2025kimi}.
However, these advancements incur substantial computational and memory costs during both training and inference. 
To address this inefficiency while preserving model expressiveness, Mixture-of-Experts (MoE)~\cite{fedus2022switch,riquelme2021scaling,lepikhin2020gshard,chi2022representation} have emerged as a promising approach. 
By sparsely activating only a subset of parameters for each token, MoE achieves significant computational efficiency without compromising performance.
This architecture has recently been extended to Vision-Language Models (VLMs)~\cite{li2024llava,ko2023meltr,achiam2023gpt,team2025kimi-vl,zhu2025internvl3,bai2025qwen2,ko2025st,lee2025vidchain}, allowing scalable multi-modal understanding while reducing computational cost.

When activating a subset of experts, most MoE architectures~\cite{liu2024deepseek,wu2024deepseek,team2025kimi,dai2024deepseekmoe,team2025kimi-vl,zhang2024clip} select the top-$K$ experts for each token at layer in a \textit{greedy} manner based on gating (or routing) scores as in Fig.~\ref{fig:teaser1}.
While simple and computationally efficient, this deterministic top-$K$ strategy restricts the \textit{exploration} of diverse expert combinations, often overlooking more optimal selections and leading the model to overfit to a small subset of experts.
To mitigate this, several studies~\cite{riquelme2021scaling,lepikhin2020gshard,zhou2022mixture,zoph2022st,chi2022representation} have proposed improved expert selection mechanisms.
For example, V-MoE~\cite{riquelme2021scaling} introduces stochasticity by adding Gaussian noise to the gating scores, yielding moderate performance gains.
However, such heuristic perturbations only partially address the exploration challenge, as they do not explicitly optimize the expert selection `policy'.
As a result, the problem of learning an optimal expert selection strategy remains largely unexplored.

To this end, we investigate MoE architectures for multi-modal understanding in VLMs and propose an expert selection policy optimization framework, \textbf{MoE-GRPO}. 
Specifically, we formulate expert selection as a sequential decision-making problem and employ a reinforcement learning (RL) algorithm, GRPO, to optimize the routing policy. 
Through MoE-GRPO, the model learns more optimal expert combinations for each token and layer by exploring diverse sampled expert sequences, guided by verifiable rewards.
During training, the model reinforces high-reward outputs while suppressing low-reward ones within each rollout group.
In addition, we introduce a \textit{modality-aware router guidance} mechanism that discourages the router from exploring experts that are infrequently activated for a given modality, thereby enabling more stable and robust policy sampling.

For evaluation, we convert the InternVL3.5-1B~\cite{wang2025internvl3} architecture into an MoE architecture and fine-tune it using MoE-GRPO.
Across a wide range of image and video understanding benchmarks, MoE-GRPO consistently outperforms standard top-$K$ routing and its variants.
Our in-depth analyses further show that MoE-GRPO encourages more diverse expert utilization and induces task-level expert specialization, leading to improved generalization across both cross-dataset evaluation and domain generalization settings.
These results indicate that the model effectively learns a routing policy that identifies effective expert combinations through reward-driven expert selection optimization.

\noindent To summarize, our contributions are \textbf{threefold}:
\begin{itemize}
    \item We propose MoE-GRPO, a novel RL-based training framework for optimizing expert selection policy in MoE-based VLMs.
    To the best of our knowledge, this is the first work to formulate expert selection as a sequential decision-making problem and optimize it through RL.
    \item We introduce a modality-aware router guidance mechanism that discourages the router from selecting experts that are rarely activated for a given modality, thereby improving training efficiency and stability.
    \item Our experiments demonstrate that MoE-GRPO outperforms standard top-$K$ routing and its variants in optimizing expert selection policies, exhibiting more diverse expert utilization and improved generalization capability.
\end{itemize}

\section{Related Works}

\noindent \textbf{Vision-Language Models (VLMs).}
Recent VLMs have been widely applied in diverse tasks for both images~\cite{li2023blip, dai2023instructblip, liu2023visual, bai2023qwen} and videos~\cite{ko2022video,zhang2023video,ko2023open,lin2024video,wang2024qwen2,wang2022internvideo} with their remarkable perception and reasoning capabilities.
Early approaches such as CLIP~\cite{radford2021learning} and ALIGN~\cite{jia2021scaling} learned joint vision-language representations through contrastive pretraining, providing strong zero-shot recognition capabilities.
Subsequent instruction-tuned models, including LLaVA~\cite{liu2023visual}, Qwen-VL~\cite{bai2023qwen}, InternVL~\cite{chen2024internvl}, and their video counterparts~\cite{zhang2023video,ko2023large,lin2024video,wang2024qwen2,wang2022internvideo,lee2025captioning,ko2025bidirectional}, advanced multi-modal understanding by pairing pretrained LLMs with powerful vision encoders and large-scale multi-modal corpora.
Despite these advances, scaling VLMs to larger model sizes or richer modalities incurs substantial computational overhead.
In this work, we adopt MoE architectures trained with RL to efficiently scale VLMs while preserving high model capacity.

\noindent \textbf{Mixture-of-Experts (MoE).}
The MoE architecture improves computational efficiency by activating only a subset of parameters within large pretrained models.
Early MoE frameworks such as GShard~\cite{lepikhin2020gshard} and Switch Transformer~\cite{fedus2022switch} demonstrated that replacing the feed-forward networks in Transformers with MoE layers enables pretraining at the trillion-parameter scale.
Subsequent research has advanced MoE models along three main directions: (1) designing more effective expert architectures~\cite{dai2024deepseekmoe, wu2024deepseek}, (2) developing improved routing algorithms~\cite{riquelme2021scaling, zhou2022mixture}, and (3) enhancing load balancing across experts~\cite{zoph2022st, chi2022representation}.
However, most existing approaches deterministically select the top-$K$ experts for each token, which restricts diverse expert utilization and results in overfitting to a small subset of experts.
To address this limitation, we aim to search for an optimal expert routing policy using RL beyond deterministic top-$K$ selection.

\noindent \textbf{Reinforcement learning with verifiable rewards.}
Reinforcement learning (RL)~\cite{schulman2017proximal, ouyang2022training, rafailov2023direct, shao2024deepseekmath}-based fine-tuning methods, such as RLHF~\cite{ouyang2022training} and DPO~\cite{rafailov2023direct}, have been widely adopted to align LLM outputs with human preferences.
Recently, DeepSeek~\cite{shao2024deepseekmath} introduced Group Relative Policy Optimization (GRPO), which first generates multiple chain-of-thought (CoT) responses and then optimizes the LLM using outcome-based verifiable rewards derived from these generations.
GRPO and its variants~\cite{zhang2025r1,zheng2025group,yu2025dapo,park2025deepvideo} have improved LLM reasoning performance by exploring and verifying multiple output sequences, outperforming traditional supervised fine-tuning methods that rely solely on ground-truth data.
Building upon these successes, several studies have extended GRPO to VLMs.
For instance, Video-R1~\cite{feng2025video} introduces video-specific reward functions to train VLMs via GRPO.
Motivated by the success of GRPO, we formulate expert routing at each layer as a sequence of actions within the policy and propose MoE-GRPO, an RL-based framework that optimizes expert selection in MoE-based VLMs through RL.
\begin{figure*}[t]
    \centering
    \includegraphics[width=\linewidth]{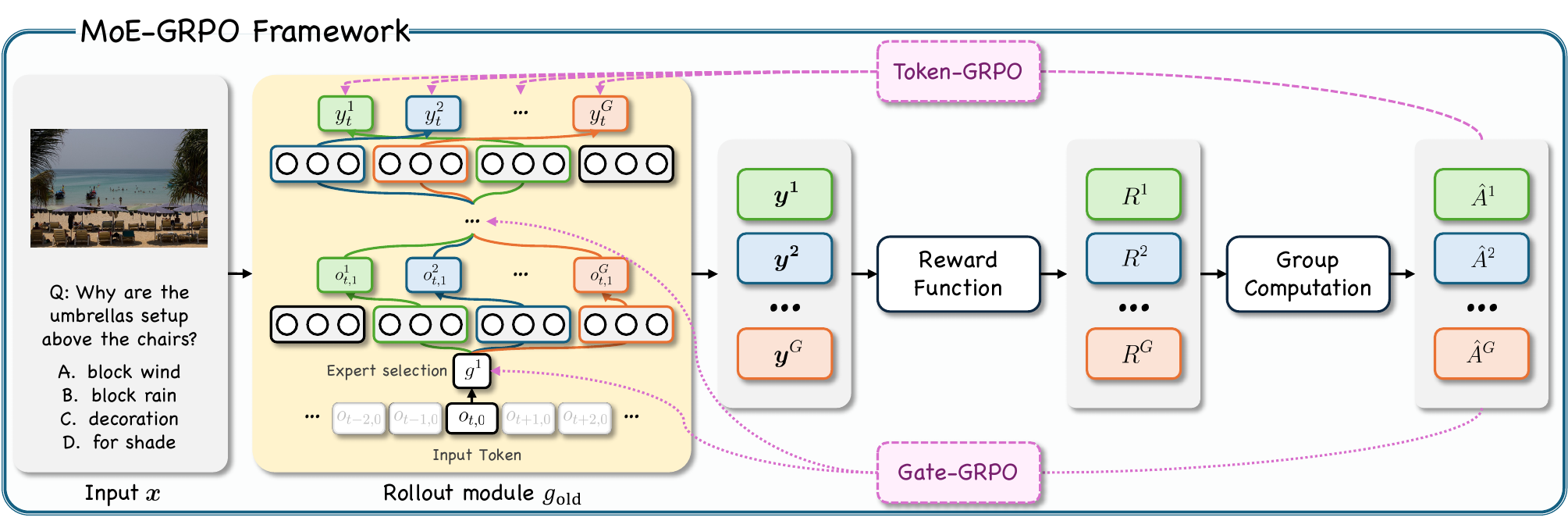}
    \caption{\textbf{Overall pipeline of MoE-GRPO.}
        Given an input image (or video) and a question, denoted as $\boldsymbol{x}$, the rollout module $g_\text{old}$ samples $G$ expert routing policies, \ie, $\{\boldsymbol{E}^i\}_{i=1}^G \sim g_\text{old}(\boldsymbol{E}|\boldsymbol{x})$, where each policy $\boldsymbol{E}^i$ represents a sequence of expert selections across layers. 
        Under each rollout $\boldsymbol{E}^i$, the model generates an output token sequence $\boldsymbol{y}^i$, and a corresponding reward $R^i$ is computed by the reward function. 
        The relative reward of each rollout is evaluated within its group to derive the advantage value $\hat{A}^i$, which guides the policy update toward higher-reward expert combinations.
        To jointly optimize token-level generation and layer-wise expert routing, the overall training objective of MoE-GRPO consists of two sub-objectives: \textit{Token-GRPO}, which optimizes token-level generation quality, and \textit{Gate-GRPO}, which refines layer-wise expert selection through the gating network.
    }
    \label{fig:main}
\end{figure*}
\section{Method}

Standard MoE architectures sparsely activate a small subset of experts by deterministically selecting the top-$K$ experts for each token. 
However, this strategy restricts the exploration of diverse expert combinations, potentially overlooking more optimal routing policies. 
To address this issue, we formulate expert selection as a sequential decision-making problem and propose MoE-GRPO, an RL-based framework for optimizing expert routing. 
Leveraging Group Relative Policy Optimization (GRPO), the model stochastically explores expert assignments and exploits those that yield higher rewards for each token. 
Moreover, to enable efficient and stable policy optimization in multi-modal MoE architectures, we introduce a \textit{modality-aware router guidance} mechanisim that discourages the router from selecting experts that are rarely activated for a given modality.
The overall pipeline of MoE-GRPO is illustrated in Fig.~\ref{fig:main}.
Before introducing MoE-GRPO, we first summarize the basic concepts of MoE and GRPO.

\subsection{Preliminaries}

\noindent \textbf{Mixture-of-Experts~(MoE).}
We formalize the standard MoE within the Transformer architecture~\cite{vaswani2017attention}.
At each layer, every token is routed among $N$ experts, where the $k$-th expert at layer $l$ is denoted as $e_{k, l}$ and implemented as a feed-forward network (FFN).
Specifically, given the hidden representation $h_{t, l}$ of the $t$-th token  after the self-attention block of layer $l$, a gating network $g^l$ computes gating scores $g^l(h_{t,l}) \in \mathbb{R}^N$, which represent the assignment probabilities over  the $N$ experts as:
\begin{equation}
    g^l(h_{t,l}) = \text{softmax}(\text{linear}(h_{t, l})) \in \mathbb{R}^N.
\end{equation}
The standard MoE layer then selects the top-$K$ experts according to $g^l(h_{t,l})$ and computes the output representation $o_{t,l}$ as:
\begin{equation}
    o_{t,l} = \sum_{k \in \text{top-}K \left(g^l(h_{t,l})\right)} g^l(h_{t,l})_k \cdot e_{k,l}(h_{t,l}).
    \label{eq:moe}
\end{equation}
As shown in Eq.~\eqref{eq:moe}, the final output $o_{t,l}$ is calculated as a weighted sum of the selected $K$ experts, where the gating scores $g^l(h_{t,l})$ determine each expert's contribution.
This formulation highlights the deterministic top-$K$ routing mechanism that our method aims to improve through RL-based stochastic exploration.

\noindent \textbf{Group Relative Policy Optimization~(GRPO).} 
GRPO is a variant of Proximal Policy Optimization (PPO)~\cite{schulman2017proximal}, designed to enhance training efficiency through group-based relative rewards in the reinforcement fine-tuning of LLMs and VLMs.
Unlike standard PPO, which relies on a single-sample value function estimate, GRPO approximates the value function using the average reward over multiple sampled trajectories (\ie, rollouts).
Given an input prompt $\boldsymbol{x}$, $G$ rollout sequences $\{\boldsymbol{y}^{i}\}_{i=1}^G$ are sampled from the old policy $\pi_\text{old}$.
Then, GRPO optimizes the following objective:
\begin{equation}
\begin{split}
    & \mathcal{L}_{\text{GRPO}} = \mathbb{E}_{\boldsymbol{x} \sim \mathcal{D}, \{\boldsymbol{y}^i\}_{i=1}^G \sim \pi_\text{old}(\boldsymbol{y}|\boldsymbol{x})} \\
    &-\frac{1}{\left\lvert \boldsymbol{y}^i \right\rvert}\sum_{t=1}^{\left\lvert \boldsymbol{y}^i \right\rvert}  \min \left[r_t^i  \hat{A}^i, 
    \text{clip} \left(r_t^i, 1-\epsilon, 1+\epsilon\right)\hat{A}^i \right],\\
    &\text{where }\: r_t^i = \frac{\pi_\theta \left(y_t^i | \boldsymbol{x}, \boldsymbol{y}_{<t}^i \right)}{\pi_\text{old}\left(y_t^i | \boldsymbol{x}, \boldsymbol{y}_{<t}^i \right)},
    \label{eq:grpo}
\end{split}
\end{equation}
and $\epsilon$ denotes the hyperparameter for the clipping function.
Here, $\hat{A}^i$ denotes the normalized advantage of the $i$-th rollout, computed as $\hat{A}^i = \frac{R^i - \text{mean}\left(\{R^j\}_{j=1}^G\right)}{\text{std}\left(\{R^j\}_{j=1}^G\right)}$ and $R^i$ represents the reward of the output sequence $\boldsymbol{y}^i$ calculated by a predefined reward function.
By Eq.~\eqref{eq:grpo}, GRPO adaptively updates the policy using relative rewards within each rollout group, enabling the model to reinforce high-reward actions while discouraging low-reward ones.
\subsection{MoE-GRPO}

In standard GRPO, an action is defined as sampling the next output token, whereas in MoE-GRPO, an action corresponds to selecting the top-$K$ experts for a given token at a specific layer.
Accordingly, while the action space of GRPO is restricted to the generated token sequence $[y_1, y_2, \dots, y_T]$, limiting optimization to the output level, MoE-GRPO expands the action space to include expert routing decisions across both tokens and layers, \ie, $[o_{1,1}, o_{1,2}, \dots, o_{2,1}, o_{2,2}, \dots o_{T,L}]$. 
This broader formulation provides fine-grained, hierarchical control over the model's behavior, enabling the policy to jointly optimize token-level generation and layer-wise expert selection through the gating network.
Therefore, the overall training objective of MoE-GRPO consists of two sub-objectives: \textit{Token-GRPO} to improve sequence generation and \textit{Gate-GRPO} to enhance the expert routing decision process.

First, the objective of Token-GRPO is defined as:
\begin{equation}
\begin{split}
    & \mathcal{L}_{\text{Token-GRPO}} = \mathbb{E}_{\boldsymbol{x} \sim \mathcal{D}, \{\boldsymbol{E}^i\}_{i=1}^G \sim g_\text{old}(\boldsymbol{E}|\boldsymbol{x})
    ,\boldsymbol{y}^i \sim  \pi_\text{old} \left(\boldsymbol{y}|\boldsymbol{x};\boldsymbol{E}^i \right)} \\
    & -\frac{1}{\left\lvert \boldsymbol{y}^i \right\rvert}\sum_{t=1}^{\left\lvert \boldsymbol{y}^i \right\rvert} \min \left[r^i_t\hat{A}^i, 
    \text{clip} \left(r^i_t, 1-\epsilon, 1+\epsilon\right)\hat{A}^i \right],\\
    &\text{where } \: r_t^i = \frac{\pi_\theta \left(y_t^i | \boldsymbol{x}, \boldsymbol{y}_{<t}^i; \boldsymbol{E}_{<t}^i  \right)}{\pi_{{\text{old}}} \left(y_t^i | \boldsymbol{x}, \boldsymbol{y}_{<t}^i;\boldsymbol{E}_{<t}^i \right)},
    \label{eq:token-grpo}
\end{split}
\end{equation}
and $\boldsymbol{E}$ is a sequence of selected experts across all tokens and layers. $\boldsymbol{y}^i \sim \pi_\text{old} \left(\boldsymbol{y}|\boldsymbol{x};\boldsymbol{E}^i \right)$ denotes the token sequence generated under the $i$-th sampled set of experts rollout $\boldsymbol{E}^i$.
In Eq.~\eqref{eq:token-grpo}, a sequence of gating networks $g_\text{old}$ samples $G$ rollout trajectories of expert assignments. 
The model then learns to reinforce expert selection policies that yield high-reward token generations while suppressing those associated with low rewards, based on relative rewards within each rollout group.

In addition to token-level optimization with Token-GRPO, we introduce Gate-GRPO, designed to refine layer-wise expert routing by optimizing the expert selection policy of each gate network $g^{l}$.
Gate-GRPO is defined as:
\begin{equation}
\begin{split}
    & \mathcal{L}_{\text{Gate-GRPO}} = \mathbb{E}_{\boldsymbol{x} \sim \mathcal{D}, \{\boldsymbol{E}^i\}_{i=1}^G \sim g_{ {\text{old}}}(\boldsymbol{E}|\boldsymbol{x})
    , \boldsymbol{y}^i \sim  \pi_\text{old} \left(\boldsymbol{y}|\boldsymbol{x};\boldsymbol{E}^i \right)}\\ -&\frac{1}{L\left\lvert \boldsymbol{y}^i \right\rvert}\sum_{l=1}^L\sum_{t=1}^{\left\lvert \boldsymbol{y}^i \right\rvert}
     \min \left[\hat{r}_{t,l}^i\hat{A}^i, 
    \text{clip} \left(\hat{r}_{t,l}^i, 1-\epsilon, 1+\epsilon\right)\hat{A}^i \right],\\
    &\text{ where } \: \hat{r}_{t,l}^i = \frac{g^l_{\theta} \left(E_{t,l}^i | \boldsymbol{x}, \boldsymbol{y}_{<t, <l}^i \right)}{g^l_{\text{old}} \left(E_{t,l}^i | \boldsymbol{x}, \boldsymbol{y}_{<t, <l}^i \right)}.
    \label{eq:gate-grpo}
\end{split}
\end{equation}
$E_{t,l}^i$ represents the set of experts selected for the $t$-th token at the $l$-th layer during the $i$-th rollout. 
Eq.~\eqref{eq:gate-grpo} encourages the gating networks to assign higher probabilities to experts, yielding high rewards, while downweighting those that lead to low rewards. 
This mechanism effectively guides each gating network toward reward-aligned expert utilization. 
Unlike Token-GRPO, which optimizes expert selection policies at the output token level, Gate-GRPO directly optimizes the gating function at each layer, providing dense and fine-grained supervision signals for the routing process.

As a result, the final objective of MoE-GRPO jointly optimizes the Token-GRPO and Gate-GRPO as:
\begin{equation}
    \mathcal{L}_{\text{MoE-GRPO}} = \mathcal{L}_{\text{Token-GRPO}} + \mathcal{L}_{\text{Gate-GRPO}}.
    \label{eq:moe-grpo}
\end{equation}
By directly applying MoE-GRPO without a supervised fine-tuning stage, the gating networks are trained from scratch, leaving no pretrained routing policy available.
Therefore, unlike conventional GRPO formulations that incorporate a KL divergence term to regularize the learned policy toward a reference model, MoE-GRPO does not rely on a reference policy.
For the reward function, we employ an accuracy-based reward that assigns a reward of 1 to correct model predictions and 0 otherwise.
MoE-GRPO converts this binary reward signal into dense supervision by propagating group-computed advantages to the router at every layer and token position, enabling direct optimization of sequential expert selection decisions.
\subsection{Modality-Aware Router Guidance}
\begin{table*}[!t]
    \centering
    \small
    \setlength{\tabcolsep}{2pt}
    \begin{adjustbox}{width=0.99\textwidth}
    \begin{tabular}{l|c c c|c c c c c|c c c c|c}
        \toprule
        & \multicolumn{3}{c|}{} & \multicolumn{5}{c|}{\textbf{Image Understanding Benchmarks}} & \multicolumn{4}{c|}{\textbf{Video Understanding Benchmarks}} \\
        \multicolumn{1}{c|}{\textbf{Models}} & \textbf{Arch.} & \textbf{\# activated} & \textbf{\# total} & \textbf{MME} & \textbf{MMBench} & \textbf{MMStar} & \textbf{MMT-Bench} & \textbf{AI2D} & \textbf{VideoMME} & \textbf{MLVU} & \textbf{LongVideoBench} & \textbf{MVBench} & \textbf{Avg.}\\
        \midrule
        \midrule
        LLaVA-OV~\cite{li2024llava} & Dense & 1B & 1B & 1,478 & 52.1 & 37.5 & - & 57.1 & 44.0 & 50.3 & 45.8 & 45.5 & - \\
        VideoChat2-Phi3~\cite{li2024mvbench} & Dense & 4B & 4B & - & - & - & - & - & - & - & - & 55.1 & - \\
        Mini-InternVL1.5~\cite{gao2024mini} & Dense & 2B & 2B & 1,899 & 70.9 & - & - & 69.8 & 42.9 & - & - & 37.0 & - \\
        InternVL2~\cite{chen2024far} & Dense & 1B & 1B & 1,794 & 65.4 & 45.7 & 49.5 & 64.1 & 42.6 & 51.6 & 43.3 & 57.5 & - \\
        InternVL2.5~\cite{chen2024expanding} & Dense & 1B & 1B & 1,950 & 70.7 & 50.1 & 50.3 & 69.3 & 50.3 & 57.3 & 47.9 & 64.3 & - \\
        DeepVideo-R1~\cite{park2025deepvideo} & Dense & 3B & 3B & - & - & - & - & - & 51.1 & - & - & 49.6 & - \\
        LLaVA-NeXT-Video~\cite{zhang2024llavanext-video} & Dense & 7B & 7B & - & - & - & - & - & - & - & - & 46.5 & - \\
        VideoChat2~\cite{li2024mvbench} & Dense & 7B & 7B & - & - & - & - & - & 33.7 & - & - & 51.1 & - \\ 
        VideoLLaMA2~\cite{cheng2024videollama} & Dense & 7B & 7B & - & - & - & - & - & 45.1 & - & - & 53.4 & - \\
        LongVA~\cite{zhang2024long} & Dense & 7B & 7B & - & - & - & - & 70.7 & 47.9 & - & - & - & - \\
        \midrule
        InternVL3.5 + Det-FT & MoE & 1.3B & 2.9B & 1,660 & 75.8 & 45.6 & 51.8 & 62.7 & 45.6 & 48.6 & 45.3 & 56.7 & 54.0 \\
        InternVL3.5 + Stoch-FT-Multi & MoE & 1.3B & 2.9B & 1,458 & 73.9 & 43.3 & 51.2 & 61.8 & 45.9 & 50.3 & \textbf{47.0} & 56.5 & 53.7 \\
        InternVL3.5 + Stoch-FT-Noise & MoE & 1.3B & 2.9B & 1,684 & 76.3 & \textbf{46.1} & 52.0 & 62.4 & 45.1 & 51.1 & 45.3 & 55.8 & 54.3 \\
        \midrule
        \rowcolor[HTML]{F0F8FF}
        InternVL3.5 + MoE-GRPO (\textbf{ours}) & MoE & 1.3B & 2.9B & \textbf{1,693} & \textbf{77.5} & 45.7 & \textbf{54.8} & \textbf{65.8} & \textbf{46.6} & \textbf{53.1} & 46.5 & \textbf{58.4} & \textbf{56.0} \\
        \bottomrule
    \end{tabular}
    \end{adjustbox}
    \caption{\textbf{Results on multi-modal understanding benchmarks.}
        \# activated and \# total denote the number of activated and total parameters.
        The last column reports the average accuracy across all benchmarks, excluding MME.
    }
    \label{tab:main}
\end{table*}

In RL-based training, the router must explore a large search space of sequential expert selections, often requiring long training schedules to sufficiently cover this space. 
This motivates the need for structured guidance that reduces unnecessary exploration and improves training efficiency. 
To this end, we introduce a modality-aware router guidance that discourages the router from exploring experts that are infrequently activated for a given modality.
For example, when processing visual tokens, the policy prioritizes experts frequently activated for visual representations while de-emphasizing those primarily associated with textual inputs, and vice versa.

To quantify modality specialization of each expert, we define modality-awareness scores, a vision-awareness score $\hat{s}_v(e_i)$ and a text-awareness score $\hat{s}_t(e_i)$, that estimate each expert's affinity for visual or textual inputs. 
Specifically, we first compute $N_v\left(e_i\right)$ and $ N_t\left(e_i \right)$, the number of times expert $e_i$ is selected by vision and text tokens, respectively, baed on top-$K$ routing.
Then, the modality-awareness scores, $\hat{s}_v(e_i)$ and $\hat{s}_t(e_i)$, are computed as:
\begin{equation}
    \begin{split}
        s_v(e_i) = \frac{N_v(e_i)}{\sum_j N_v(e_j)}, &\;\; s_t(e_i) = \frac{N_t(e_i)}{\sum_j N_t(e_j)} \\
        \hat{s}_v(e_i) = \frac{s_v(e_i)}{s_v(e_i) + s_t(e_i)}, &\;\; \hat{s}_t(e_i) = \frac{s_t(e_i)}{s_v(e_i) + s_t(e_i)},
    \end{split}
    \label{eq:score}
\end{equation}
Based on these scores, we deactivate the bottom $P\%$ of the experts by setting their gating scores to $-\infty$, thereby constraining exploration to modality-relevant experts.
Within the remaining search space, the gating network $g_\text{old}$ samples $K$ experts according to the adjusted gating scores via multinomial sampling, yielding $G$ stochastic rollout policies.
This strategy ensures that the router primarily explores modality-consistent experts while maintaining sufficient diversity for robust and efficient policy learning.
\section{Experiments}
\begin{table*}[!t]
    \centering
    \setlength{\tabcolsep}{2pt}
    \begin{adjustbox}{width=0.99\textwidth}
    \begin{tabular}{l|c c c c c c c c c c|c}
        \toprule
        \multicolumn{1}{c|}{\textbf{Models}} & \textbf{Caltech101} & \textbf{OxfordPets} & \textbf{StanfordCars} & \textbf{Flowers102} & \textbf{Food101} & \textbf{Aircraft} & \textbf{SUN397} & \textbf{DTD} & \textbf{EuroSAT} & \textbf{UCF101} & \textbf{Avg.} \\
        \midrule
        \midrule
        CoOp~\cite{zhou2022learning} & 93.7 & 89.1 & 64.5 & 68.7 & 85.3 & 18.5 & 64.2 & 41.9 & 46.4 & 66.6 & 63.9 \\
        CoCoOp~\cite{zhou2022conditional} & 94.4 & 90.1 & 65.3 & 71.9 & 86.1 & 22.9 & 67.4 & 45.7 & 45.4 & 68.2 & 65.7 \\
        MaPLe~\cite{khattak2023maple} & 93.5 & 90.5 & 65.6 & 72.2 & 86.2 & 24.7 & 67.0 & 46.5 & 48.1 & 68.7 & 66.3 \\
        \midrule
        CLIP-MoE~\cite{zhang2024clip} & 95.8 & \textbf{92.3} & 74.9 & 72.1 & 88.7 & 29.0 & 70.1 & \textbf{54.9} & 57.6 & 73.0 & 70.8 \\
        CLIP-MoE + Det-FT & 94.1 & 89.7 & 66.1 & 68.6 & 88.3 & 25.1 & 67.0 & 53.1 & \textbf{62.2} & 73.9 & 68.8 \\
        \rowcolor[HTML]{F0F8FF}
        CLIP-MoE + MoE-GRPO (\textbf{ours}) & \textbf{95.9} & 91.6 & \textbf{75.5} & \textbf{75.9} & \textbf{90.5} & \textbf{30.0} & \textbf{71.1} & 53.8 & 58.9 & \textbf{76.2} & \textbf{71.9} \\
        \bottomrule
    \end{tabular}
    \end{adjustbox}
    \caption{\textbf{Results on cross-dataset evaluation.}
        We train the model on the source ImageNet dataset for three epochs under the 16-shot setting and evaluate it on 10 target datasets.
    }
    \label{tab:main-clip}
\end{table*}

To apply MoE-GRPO, we convert the InternVL3.5-1B~\cite{wang2025internvl3} into an MoE architecture by activating $K=2$ experts from a total of $N=8$ experts at each layer.
As a result, only 1.3B parameters are activated out of 2.9B total parameters.
To further demonstrate the generalization capability of models trained with MoE-GRPO, we conduct cross-dataset evaluation and domain generalization experiments on image classification using CLIP-MoE~\cite{zhang2024clip}, where 0.7B parameters are activated out of 1.2B total parameters by selecting two experts from four at each layer.
We first compare MoE-GRPO with both dense and MoE models in Sec.~\ref{subsec:exp:main}, followed by ablation studies and in-depth analyses in Sec.~\ref{subsec:exp:ablation} and Sec.~\ref{subsec:exp:analyses}, respectively.

\noindent \textbf{Implementation details.}
In MoE-GRPO, we set the number of rollouts to $G = 8$ and apply multinomial sampling within each gating network.
For modality-aware router guidance, we deactivate the bottom 25\% of experts for each modality. 
During rollout, we employ greedy decoding for language generation and introduce stochasticity only in the expert selection process.
For training, we sample 100K multi-choice visual instruction-tuning examples from OneThinker~\cite{feng2025onethinker} and conduct 25K training steps on 4 $\times$ RTX Pro 6000 Blackwell Max-Q GPUs with a batch size of 4.
The learning rate is set to 1e-6 with cosine scheduling, and the total training time is approximately one day.
We apply the load-balancing loss introduced in Switch Transformers~\cite{fedus2022switch}. 
We use up to 8 images for image inputs and 8 frames for video inputs.
During inference, expert selection is performed deterministically by choosing the top-$K$ experts.

\noindent \textbf{Baselines.}
To evaluate the effectiveness of MoE-GRPO, we compare it against standard fine-tuning strategies and their variants for MoE architectures:
\begin{itemize}
    \item \textbf{Deterministic Fine-Tuning (\textit{Det-FT})} selects the top-$K$ experts according to the gating scores.
    \item \textbf{Stochastic Fine-Tuning with Multinomial Sampling (\textit{Stoch-FT-Multi})} samples $K$ experts from a multinomial distribution parameterized by the gating scores, following V-MoE~\cite{riquelme2021scaling}.
    \item \textbf{Stochastic Fine-Tuning with Gaussian Noise (\textit{Stoch-FT-Noise})} perturbs the gating scores with Gaussian noise prior to selecting the top-$K$ experts, following~\cite{zhu2024mote}.
\end{itemize}
To ensure a fair comparison, we apply the load-balancing loss to all baselines, consistent with MoE-GRPO.

\subsection{Main Results}
\label{subsec:exp:main}

\noindent \textbf{Results on multi-modal understanding benchmarks.}
We compare MoE-GRPO with three fine-tuning baselines for MoE architectures (Det-FT, Stoch-FT-Multi, and Stoch-FT-Noise) on multi-modal understanding benchmarks in Tab.~\ref{tab:main}. 
Comparisons between Det-FT, Stoch-FT-Multi, and Stoch-FT-Noise indicate that merely introducing stochasticity does not consistently improve performance. 
Although stochastic routing increases exploration over experts, these methods do not explicitly optimize the expert selection policy, limiting their effectiveness.
In contrast, MoE-GRPO directly optimizes the routing policy via RL, resulting in consistent performance improvements over all fine-tuning baselines on 7 out of 9 evaluated benchmarks.
Consequently, it surpasses the three baselines by 2.0\%, 2.3\%, and 1.7\% in terms of average accuracy, respectively.
Overall, these results demonstrate that RL-based direct expert selection policy optimization improves generalization and leads to more robust performance across diverse multi-modal image and video understanding tasks.

\noindent \textbf{Results of cross-dataset generalization.}
We further apply MoE-GRPO to CLIP-MoE~\cite{zhang2024clip}, a vision encoder based on a MoE architecture, to evaluate its generalization capability.
Specifically, we train CLIP-MoE + MoE-GRPO for three epochs on the source ImageNet dataset~\cite{deng2009imagenet} under the 16-shot setting and evaluate the model on 10 target datasets. 
As shown in Tab.~\ref{tab:main-clip}, training with Det-FT leads to performance degradation compared to the baseline CLIP-MoE, indicating a loss of generalization due to overfitting.
In contrast, MoE-GRPO outperforms Det-FT on 9 out of 10 evaluation datasets, achieving an average accuracy gain of 3.1\%. 

\begin{table}[!t]
    \centering
    \begin{adjustbox}{width=\linewidth}
    \begin{tabular}{l|c|c c c c|c}
        \toprule
        \multicolumn{1}{c|}{\textbf{Models}} & \textbf{ImageNet} & \textbf{ImageNetV2} & \textbf{ImageNet-S} & \textbf{ImageNet-A} & \textbf{ImageNet-R} & \textbf{Avg.} \\ 
        \midrule
        \midrule
        CLIP~\cite{radford2021learning} & 66.7 & 60.8 & 46.2 & 47.8 & 74.0 & 59.1 \\
        CoOP~\cite{zhou2022learning} & 71.5 & 64.2 & 48.0 & 49.7 & 75.2 & 61.7 \\
        CoCoOp~\cite{zhou2022conditional} & 71.0 & 64.1 & 48.8 & 50.6 & 76.2 & 62.1 \\
        MaPLe~\cite{khattak2023maple} & 70.7 & 64.1 & 49.2 & 50.9 & 77.0 & 62.4 \\
        \midrule
        CLIP-MoE~\cite{zhang2024clip} & 72.9 & 72.6 & 57.4 & 45.6 & 68.7 & 63.4 \\
        + Det-FT & 76.1 & 75.0 & 57.0 & 49.1 & 72.7 & 66.0 \\
        \rowcolor[HTML]{F0F8FF}
        + MoE-GRPO (\textbf{ours}) & \textbf{77.6} & \textbf{76.9} & \textbf{59.0} & \textbf{50.3} & \textbf{73.6} & \textbf{67.5} \\ 
        \bottomrule
    \end{tabular}
    \end{adjustbox}
    \caption{\textbf{Results of domain generalization.}
        We train the model on the source ImageNet dataset for three epochs under the 16-shot setting and evaluate it on four out-of-domain target datasets.
    }
    \label{tab:imagenet}
\end{table}
\noindent \textbf{Results of domain generalization.}
We evaluate the out-of-domain generalization of MoE-GRPO by assessing the transferability of the ImageNet-trained model to four out-of-domain datasets in Tab.~\ref{tab:imagenet}. 
While both Det-FT and MoE-GRPO substantially improve performance on the source ImageNet dataset, Det-FT often degrades accuracy on out-of-domain datasets. 
For instance, on ImageNet-S, the accuracy decreases by 0.4\% compared to CLIP-MoE. 
In contrast, training with MoE-GRPO consistently outperforms both CLIP-MoE and CLIP-MoE + Det-FT with average gains of 4.1\% and 1.5\%, respectively.
These results demonstrate that RL-based expert routing mitigates overfitting and enhances cross-domain generalization by promoting diverse expert utilization and effective use of model capacity.
\subsection{Ablation Studies}
\label{subsec:exp:ablation}

\begin{table}[!t]
    \centering
    \begin{adjustbox}{width=\linewidth}
    \begin{tabular}{c c|c c|c c|c}
        \toprule
        \textbf{Token-GRPO} & \textbf{Gate-GRPO} & \textbf{MMBench} & \textbf{MMStar} & \textbf{MLVU} & \textbf{LongVideoBench} & \textbf{Avg.} \\ 
        \midrule
        \midrule
        \ding{52} & - & 74.8 & 44.4 & 51.2 & 45.0 & 53.9 \\
        - & \ding{52} & 72.3 & 39.9 & 47.5 & 43.8 & 50.9 \\
        \midrule
        \rowcolor[HTML]{F0F8FF}
        \ding{52} & \ding{52} & \textbf{77.5} & \textbf{45.7} & \textbf{53.1} & \textbf{46.5} & \textbf{55.7} \\
        \bottomrule
    \end{tabular}
    \end{adjustbox}
    \caption{\textbf{Ablation studies on MoE-GRPO.}
    }
    \label{tab:moe-grpo}
\end{table}
\noindent \textbf{Ablation studies on MoE-GRPO.}
Tab.~\ref{tab:moe-grpo} presents the individual contributions of Token-GRPO and Gate-GRPO. 
Applying Gate-GRPO alone leads the average accuracy to drop substantially from 55.7\% to 50.9\%, indicating that optimizing expert routing without explicitly shaping token-level generation fails to align the model with task-level rewards. 
This finding suggests that token-level policy optimization is more directly coupled with reward improvement and is therefore indispensable.
In contrast, adopting only Token-GRPO results in a 1.8\% decrease in average accuracy, implying that while token-level optimization captures the primary reward signal, it does not explicitly regularize layer-wise expert routing. 
Gate-GRPO complements this limitation by providing dense supervisory signals to the routing modules at each layer, thereby facilitating more effective expert selection. 
Overall, these results demonstrate that Token-GRPO and Gate-GRPO are both necessary and complementary components of MoE-GRPO.

\begin{table}[!t]
    \centering
    \setlength{\tabcolsep}{2pt}
    \begin{adjustbox}{width=\linewidth}
    \begin{tabular}{l c c|c c|c}
        \toprule
        \multicolumn{1}{c|}{\textbf{Router guidance }} & \textbf{MMBench} & \textbf{MMStar} & \textbf{MLVU} & \textbf{LongVideoBench} & \textbf{Avg.} \\ 
        \midrule
        \midrule
        MoE-GRPO \\
        \multicolumn{1}{l|}{+ modality-agnostic (noise)} & 76.3 & 45.3 & 49.5 & 45.6 & 54.2 \\
        \multicolumn{1}{l|}{+ modality-agnostic (multi.)} & 76.7 & \textbf{46.1} & 50.8 & 45.6 & 54.8 \\
        \midrule
        \rowcolor[HTML]{F0F8FF}
        \multicolumn{1}{l|}{+ modality-aware (\textbf{ours})} & \textbf{77.5} & 45.7 & \textbf{53.1} & \textbf{46.5} & \textbf{55.7} \\
        \bottomrule
    \end{tabular}
    \end{adjustbox}
    \caption{\textbf{Ablation studies on modality-aware router guidance.}
        We compare modality-aware router guidance with two modality-agnostic expert selection mechanisms, Gaussian noise and multinomial sampling.
    }
    \label{tab:exploration}
\end{table}
\noindent \textbf{Ablation studies on modality-aware router guidance.}
Tab.~\ref{tab:exploration} presents ablation studies on modality-aware router guidance. 
While our proposed modality-aware router guidance explicitly discourages the router from exploring experts that are infrequently activated for a given modality, we construct two modality-agnostic exploration baselines that do not condition on modality information: (1) perturbing the gating scores with Gaussian noise (noise), and (2) sampling experts via multinomial sampling (multi.) during rollout. 
As shown in Tab.~\ref{tab:exploration}, modality-aware router guidance outperforms both the Gaussian-noise and multinomial exploration baselines by 1.5\% and 0.9\%, respectively.

\begin{figure}[t]
    \begin{subfigure}[t]{0.49\linewidth}
         \centering
         \includegraphics[width=\linewidth]{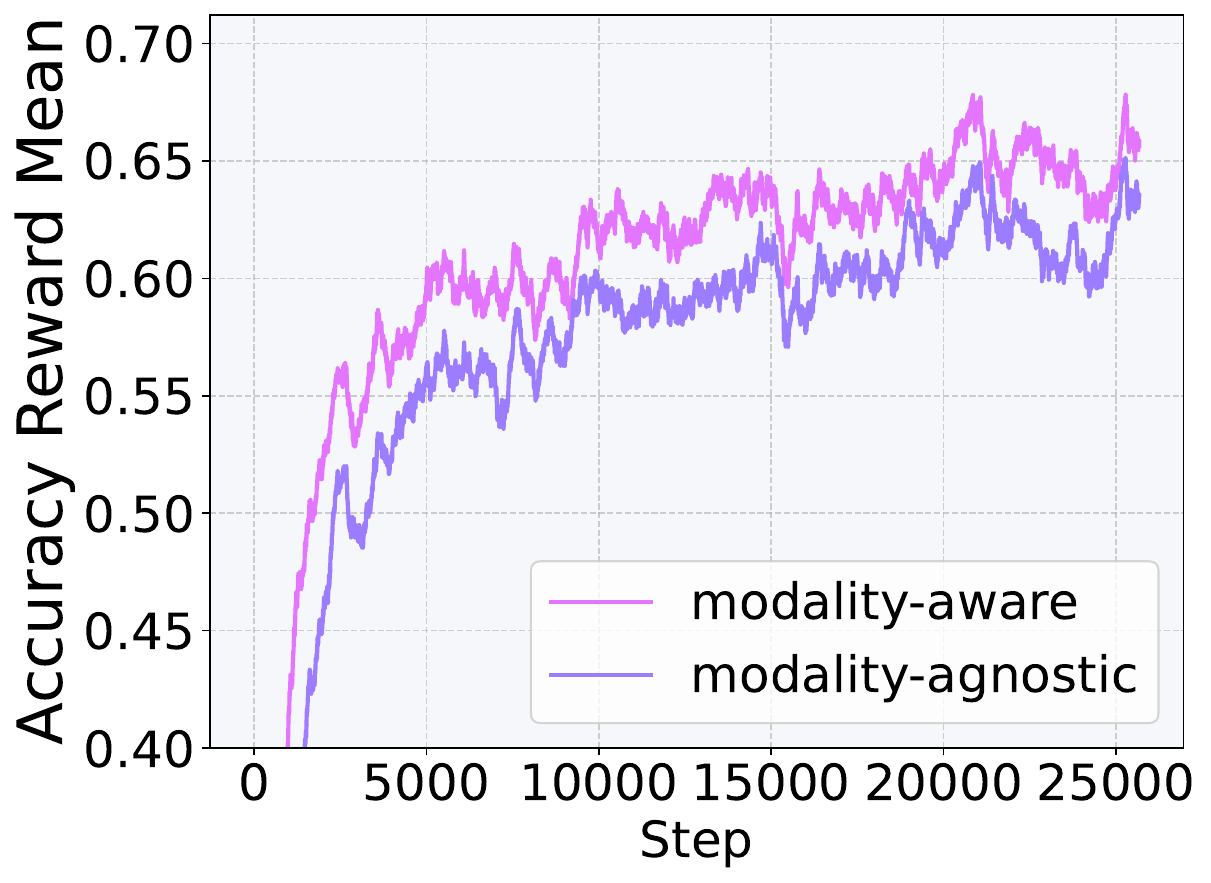}
         \caption{Reward mean.}
         \label{fig:mean}
    \end{subfigure}
    \begin{subfigure}[t]{0.49\linewidth}
         \centering
         \includegraphics[width=\linewidth]{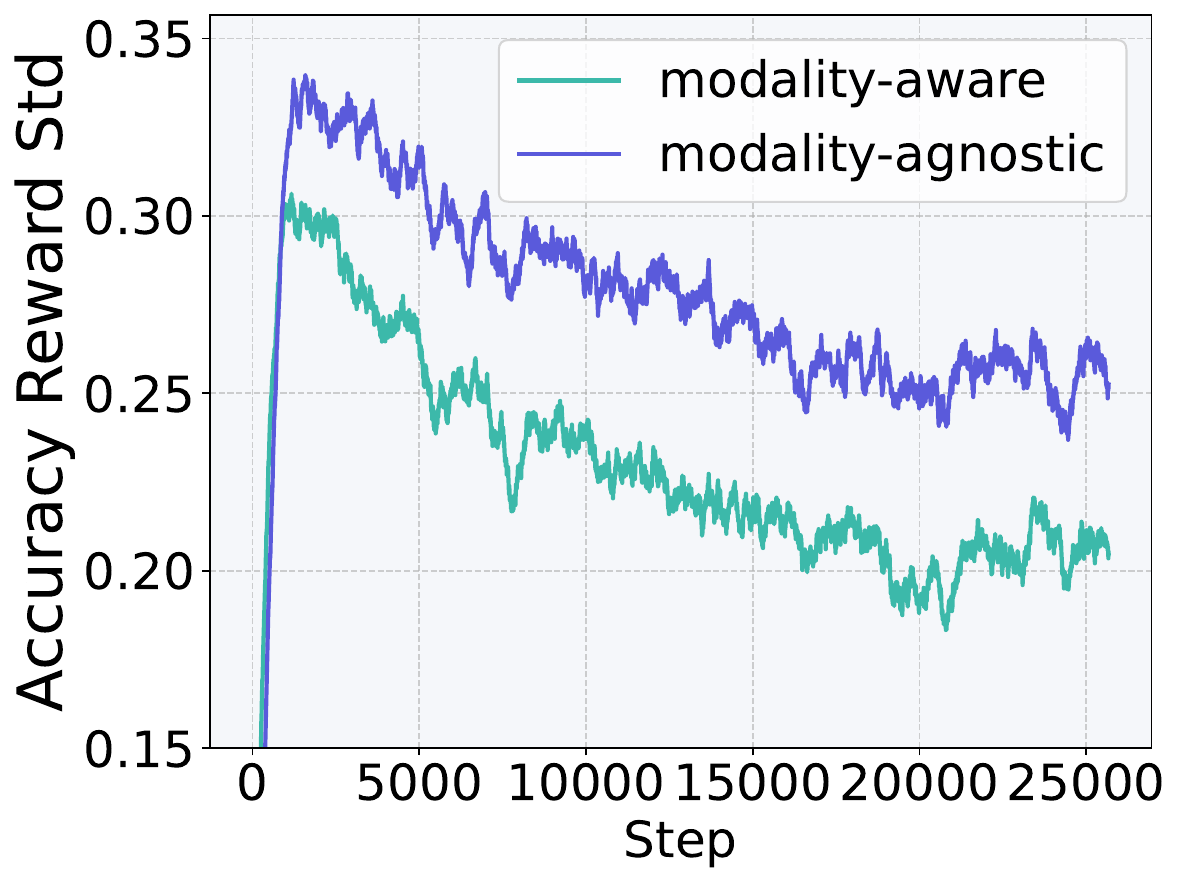}
         \caption{Reward standard deviation.}
         \label{fig:std}
    \end{subfigure}
    \caption{\textbf{Training curves.}
        (a) and (b) present the mean and standard deviation of the accuracy reward of MoE-GRPO, comparing our modality-aware router guidance with the modality-agnostic (multi.) expert selection baseline.
    }
    \label{fig:reward}
\end{figure}
Fig.~\ref{fig:reward} presents the training curves of MoE-GRPO under both modality-aware router guidance and modality-agnostic (multi.) expert selection, illustrating the mean and standard deviation of the accuracy reward. 
As shown in Fig.~\ref{fig:mean}, the mean accuracy steadily increases for both methods, indicating that the model progressively learns to generate correct answers through RL-driven expert selection. 
Fig.~\ref{fig:std} further illustrates a gradual reduction in the standard deviation of accuracy within each rollout group, reflecting increased policy stability over time. 
Notably, modality-aware router guidance converges more rapidly in terms of mean reward and exhibits lower reward variance than modality-agnostic (multi.) expert selection. 
This improvement arises from avoiding unnecessary exploration of irrelevant experts and more effectively leveraging modality-specific expert patterns. 
These results underscore the importance of modality-aware routing, which guides the router toward modality-relevant experts while avoiding unnecessary exploration, thereby enabling faster and more stable convergence in policy learning for vision-language models.

\begin{table}[!t]
    \centering
    \begin{adjustbox}{width=\linewidth}
    \begin{tabular}{c|c c|c c|c}
        \toprule
        \textbf{RL Methods} & \textbf{MMBench} & \textbf{MMStar} & \textbf{MLVU} & \textbf{LongVideoBench} & \textbf{Avg.} \\ 
        \midrule
        \midrule
        DAPO~\cite{yu2025dapo} & 75.0 & \textbf{46.7} & 52.0 & 47.9 & 55.0 \\
        SAPO~\cite{gao2025soft} & 76.7 & 45.7 & 50.8 & \textbf{48.5} & 55.4 \\
        \midrule
        \rowcolor[HTML]{F0F8FF}
        GRPO~\cite{shao2024deepseekmath} & \textbf{77.5} & 45.7 & \textbf{53.1} & 46.5 & \textbf{55.7} \\
        \bottomrule
    \end{tabular}
    \end{adjustbox}
    \caption{\textbf{Ablation studies on RL methods of MoE-GRPO.}
    }
    \label{tab:grpo}
\end{table}
\noindent \textbf{Ablation studies on RL methods.}
We present ablation studies of different reinforcement learning (RL) methods (GRPO~\cite{shao2024deepseekmath}, DAPO~\cite{yu2025dapo}, and SAPO~\cite{gao2025soft}), within the MoE-GRPO framework in Tab.~\ref{tab:grpo}. 
We observe that DAPO and SAPO achieve average accuracy comparable to GRPO, suggesting that expert selection policy optimization in MoE architectures is robust across different RL formulations.
\subsection{Analyses of Routing Policy}
\label{subsec:exp:analyses}

In this section, we first compare MoE-GRPO with existing routing methods and then provide an in-depth analysis of the learned routing policy of MoE-GRPO.

\begin{table}[!t]
    \centering
    \begin{adjustbox}{width=\linewidth}
    \begin{tabular}{l|c c|c c|c}
        \toprule
        \multicolumn{1}{c|}{\textbf{Routing Methods}} & \textbf{MMBench} & \textbf{MMStar} & \textbf{MLVU} & \textbf{LongVideoBench} & \textbf{Avg.} \\ 
        \midrule
        \midrule
        Expert Choice~\cite{zhou2022mixture} & 75.3 & \textbf{45.8} & 51.0 & 46.7 & 54.7 \\
        Optimal Transport~\cite{clark2022unified} & 74.4 & 45.3 & 51.0 & \textbf{47.0} & 54.4 \\
        \midrule
        MoE-GRPO w/o LB~\cite{fedus2022switch} & 77.0 & 45.4 & 52.6 & 44.0 & 54.8 \\
        \rowcolor[HTML]{F0F8FF}
        MoE-GRPO w/ LB~\cite{fedus2022switch} & \textbf{77.5} & 45.7 & \textbf{53.1} & 46.5 & \textbf{55.7} \\
        \bottomrule
    \end{tabular}
    \end{adjustbox}
    \caption{\textbf{Comparison with existing routing methods.}
        MoE-GRPO achieves superior performance compared to Expert Choice~\cite{zhou2022mixture} routing and Optimal Transport~\cite{clark2022unified} routing. 
        Moreover, it is complementary to the load-balancing (LB) objective used in Switch Transformers~\cite{fedus2022switch}, and their combination leads to further performance improvements.
    }
    \label{tab:routing}
\end{table}

\noindent \textbf{Comparison with existing routing methods.}
Tab.~\ref{tab:routing} compares MoE-GRPO with three routing methods: Expert Choice routing~\cite{zhou2022mixture}, Optimal Transport routing implemented with the Sinkhorn algorithm~\cite{clark2022unified}, and the auxiliary load-balancing (LB) loss used in Switch Transformers~\cite{fedus2022switch}.
Specifically, Expert Choice routing allows experts to select the top-$K$ tokens to enforce capacity constraints, whereas Optimal Transport routing leverages the Sinkhorn algorithm to achieve globally balanced assignments. 
In contrast, the auxiliary LB loss introduces a regularization term to the training objective to encourage balanced expert utilization. 
As in Tab.~\ref{tab:routing}, MoE-GRPO outperforms both Expert Choice and Optimal Transport routing methods by 1.0\% and 1.3\% in average accuracy, respectively.
Moreover, MoE-GRPO is complementary to the LB loss, and their combination yields further performance improvements by 0.9\%.

\begin{figure}[!t]
    \centering
    \includegraphics[width=\linewidth]{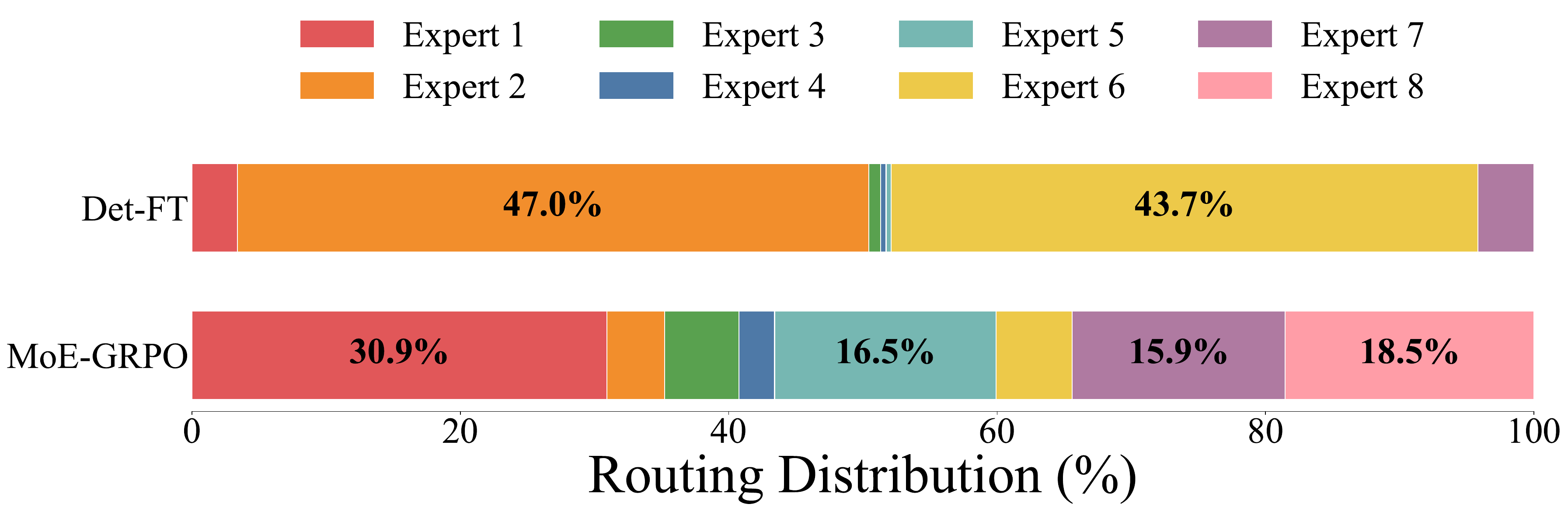}
    \caption{
        \textbf{Token-level expert utilization ratio.}
        Under MoE-GRPO, expert activation is more evenly distributed across the token sequence, resulting in more balanced expert utilization.
    }
    \label{fig:token}
\end{figure}
\begin{figure}[!t]
    \centering
    \includegraphics[width=\linewidth]{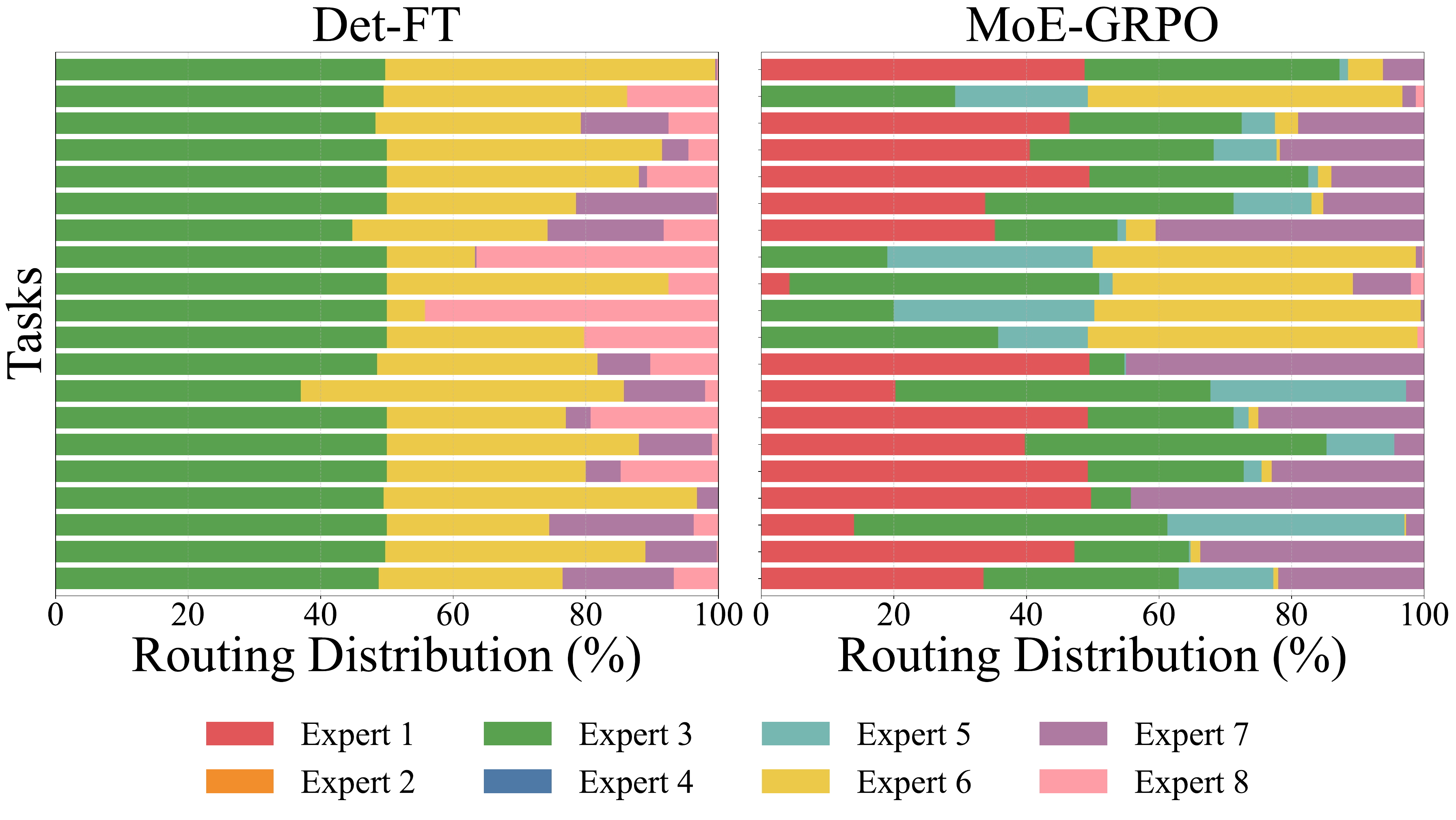}
    \caption{
        \textbf{Expert utilization ratio (x-axis) for each task (y-axis).}
        MoE-GRPO enhances task-level expert specialization by inducing more diverse expert activation patterns across tasks.
    }
    \label{fig:task}
\end{figure}
\noindent \textbf{Analysis of expert selection diversity.}
We visualize the effect of MoE-GRPO on token-level expert selection diversity over a 2K vision-language token sequence in Fig.~\ref{fig:token}. 
Det-FT predominantly activates only two experts across the sequence, whereas MoE-GRPO exhibits a substantially more diverse activation pattern, increasing the entropy of the routing distribution from 1.05 to 1.82.
In addition, Fig.~\ref{fig:task} presents the expert selection ratios across the 20 task categories in MVBench. 
Compared to Det-FT, MoE-GRPO demonstrates more distinct expert activation across tasks, indicating stronger task-level specialization. 
Consistently, the average Jensen-Shannon divergence (JSD) of task-wise expert distributions increases from 0.06 to 0.20 under MoE-GRPO.
Overall, these results suggest that MoE-GRPO enhances task-level expert specialization while maintaining balanced expert utilization at the token level.

\begin{figure}[t]
    \centering
    \includegraphics[width=\linewidth]{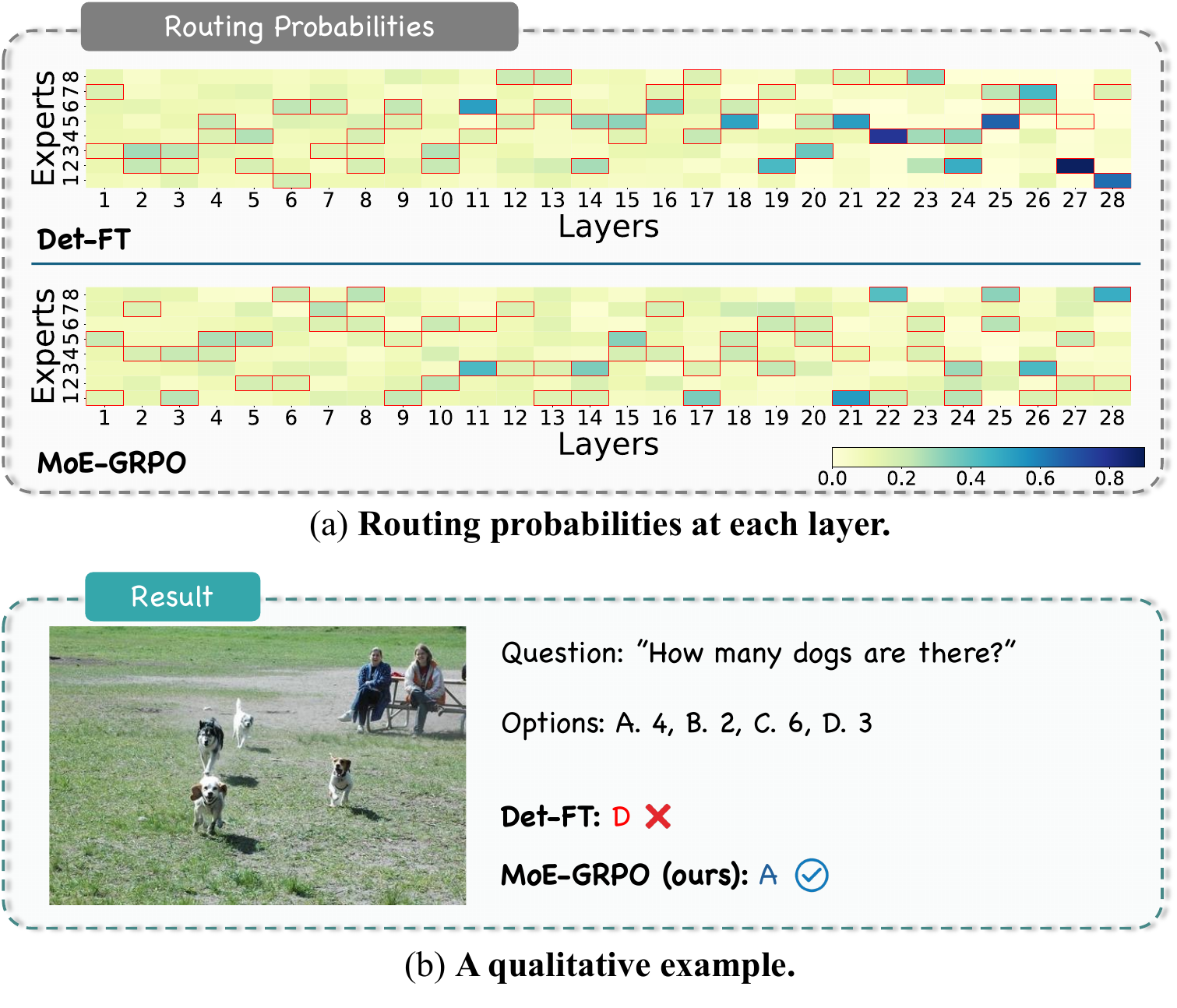}
    \caption{\textbf{A qualitative example and its routing probabilities.}
        (a) illustrates the expert routing probabilities, with the selected experts highlighted in red boxes.
        (b) presents a qualitative example demonstrating that the learned expert selection policy of MoE-GRPO yields a correct prediction, whereas the baseline Det-FT model produces an incorrect one.
    }
    \label{fig:qual}
\end{figure}
\noindent \textbf{Qualitative analysis.}
We further analyze the expert selection policy of each layer through a qualitative comparison between Det-FT and MoE-GRPO in Fig.~\ref{fig:qual}. 
As in Fig.~\ref{fig:qual}a, both methods exhibit increasingly confident routing decisions in deeper layers, as indicated by darker blue shades, suggesting progressively more decisive expert selection.
Notably, MoE-GRPO shows greater variability in routing probabilities, reflected by lighter shades of blue compared to Det-FT. 
Such diversity mitigates over-reliance on specific experts across layers.
Consequently, this adaptive routing behavior enables MoE-GRPO to produce the correct answer (`A') in Fig.~\ref{fig:qual}b, whereas Det-FT yields an incorrect prediction (`D'). 
These observations are consistent with our quantitative findings, demonstrating that MoE-GRPO fosters more diverse and adaptive expert utilization, thereby reducing expert over-specialization and improving generalization.
\section{Conclusion}

In this paper, we introduce MoE-GRPO, an RL-based framework that enables MoE-based VLMs to learn an explicit expert selection policy beyond deterministic top-$K$ routing by optimizing a reward-driven GRPO objective. 
However, unrestricted RL exploration over the sequential expert selection space can be inefficient and unstable.
To address this, we incorporate a modality-aware router guidance strategy that steers exploration toward experts most relevant to each modality, improving learning stability. 
Empirically, MoE-GRPO consistently outperforms both deterministic and stochastic routing baselines across a broad range of tasks and settings. 
Our in-depth analyses demonstrate that MoE-GRPO encourages diverse expert utilization and improved generalization by enabling more effective exploration of expert combinations during training.

\noindent \textbf{Acknowledgements.}
This work was partly supported by Institute of Information \& Communications Technology Planning \& Evaluation (IITP) grant funded by the Korea government (MSIT) (No. RS-2024-00443251 30\%, RS-2024-00457882 30\%), National Research Foundation of Korea (NRF) grant funded by the Korea government (MSIT) (NRF-2023R1A2C2005373 30\%), and Samsung Electronics Co., Ltd. (IO251218-14841-01, 10\%).

{
    \small
    \bibliographystyle{unsrt}
    \bibliography{main}
}


\end{document}